\documentclass[runningheads]{llncs}
\usepackage{amsmath}
\usepackage{multirow}
\usepackage{graphicx}
\usepackage{float}
\begin{document}
\title{Person Re-identification by analyzing Dynamic Variations in Gait Sequences\thanks{Center for Development of Advanced Computing, Kolkata}}
%
%
\author{Sandesh V Bharadwaj\inst{1,2}\orcidID{0000-0002-1491-4140} \and
Kunal Chanda\inst{2}}
\authorrunning{Sandesh Bharadwaj et. al}
%
\institute{Indian Institute of Information Technology, Design and Manufacturing, Kancheepuram, Chennai 600127, India\and
Center for Development of Advanced Computing, Kolkata 700091, India
\email{cdac.in}}
%
\maketitle              
\begin{abstract}
Gait recognition is a biometric technology that identifies individuals in a video sequence by analysing their style of walking or limb movement. However, this identification is generally sensitive to appearance changes and conventional feature descriptors such as Gait Energy Image (GEI) lose some of the dynamic information in the gait sequence. Active Energy Image (AEI) focuses more on dynamic motion changes than GEI and is more suited to deal with appearance changes. \\
\par We proposed a new approach, which allows recognizing people by analysing the dynamic motion variations and identifying people without using a database of predicted changes. In the proposed method, the active energy image is calculated by averaging the difference frames of the silhouette sequence and divided into multiple segments. Affine moment invariants are computed as gait features for each section. Next, matching weights are calculated based on the similarity between extracted features and those in the database. Finally, the subject is identified by the weighted combination of similarities in all segments. The CASIA-B Gait Database is used as the principal dataset for the experimental analysis.

\end{abstract}
\section{Introduction}
In the modern world, reliable recognition of people has become a fundamental requirement in various real-time applications such as forensics, international travel and surveillance. Biometrics have been applied to criminal identification, patient tracking in hospitals, and personalization of social services. Gait recognition is a biometric recognition technique that recognizes individuals based on their walk cycle, and has been a topic of continued interest for person identification due to the following reasons:

\begin{enumerate}
    \item First, gait recognition can be performed with low-resolution videos with relatively simple instrumentation. 
    \item Second, gait recognition can work well remotely and perform unobtrusive identification, especially under conditions of low visibility. 
    \item Third, gait biometric overcomes most of the limitations that other biometric identifiers suffer from such as face, fingerprint and iris recognition which have certain hardware requirements that add to the cost of the system. 
    \item Finally, gait features are typically difficult to impersonate or change, making them somewhat robust to appearance changes.
\end{enumerate}

Gait-based person recognition experience varying degrees of success due to internal and external factors, such as:
\begin{enumerate}
    \item Low frame rates leading to incomplete gait extraction.
    \item Incorrect or failure in identification of individuals due to
    \begin{itemize}
        \item Partial visibility due to occlusion.
        \item View variations. (lateral movement or coming at an angle)
        \item Appearance Changes. (wearing a bag/coat/jacket)
    \end{itemize}
\end{enumerate}

The rest of the paper is organized as follows. Section 2 covers related work and publications in person recognition, detailing the various methodologies implemented for gait feature extraction. Section 3 summarizes the modified approach we have implemented in our research. Section 4 explains the experimental results obtained and section 5 concludes our paper. Our main contribution is:
\begin{enumerate}
    \item Focusing on dynamic variations embedded in gait sequences using active energy images.
    \item Image segmentation of active energy images and extracting affine moment invariants as feature descriptors.
    \item Assigning dynamic weights for each segment to attribute useful features to higher priority and vice versa.
\end{enumerate}

\section{Related Work}

In the last two decades, significant efforts have been made to develop robust algorithms that can enable gait-based person recognition on real-time data. Modern gait recognition methods can be classified into two major groups, namely model-based and motion-based methods. 

In model-based methods, the human body action is described using a mathematical model and the image features are extracted by measuring the structural components of models or by the motion trajectories of the body parts. These models are streamlined based on assumptions such as pathologically normal gait.

\par Model-based systems consist of a gait sequence,a model or models, feature extractor, and a classifier. The model can be 2-dimensional or 3-dimensional, which is useful for tracking a moving person. While this method is robust to problems of occlusion, noise, scale and rotation, system effects such as viewpoint invariance and effects of physiological, psychological and environmental changes are major limitations in implementing a model-based recognition system.

\par Motion-based methods consider the human gait cycle as a sequence of images and extract binary silhouettes from these images. Motion-based approaches are not influenced by the quality of images and have the added benefit of reduced computational cost compared to model-based approaches. There are usually some post-processing steps performed on the binary images to extract static appearance features and dynamic gait features. 

\par The baseline algorithm proposed by Sarkar et al. \cite{11sarkar2005} uses silhouettes as features themselves, scaling and aligning them before use. Carley et al.\cite{Carley2019PersonRF} proposed a new biometric feature based on autocorrelation using an end-to-end trained network to capture humain gait from different viewpoints. Bobick and Davis \cite{12bobick2001} proposed the motion-energy image (MEI) and motion-history image (MHI) to convert the temporal silhouette sequence to a signal format. 
\\ \cite{kolekar2018intelligent} For a given sequence \emph{$I_t$}, MHI at pixel coordinates (\emph{x,y}) and time \emph{t} is defined as
\begin{equation}
MHI(x,y,t)_\tau = \begin{cases} \tau , \hfill I(x,y)_t = 1 \\ max(MHI(x,y,t-1)_\tau-1,0), I(x,y)_t = 0 \end{cases}
\end{equation}
where $\tau$ is a threshold value designed to capture maximum action in the image.
\\ \cite{kolekar2018intelligent} MEI is obtained by binarizing the MHI. Given a binary image sequence $J(x,y,t)$ which contains regions of motion, MEI is defined as follows
\begin{equation}
    MEI(x,y,t) = \bigcup_{i=0}^{\tau-1} J(x,y,t-i)
\end{equation}

\par Han and Bhanu \cite{13han2006} used the idea of MEI to propose the Gait Energy Image (GEI) for individual recognition. GEI converts the spatio-temporal information of one walking cycle into a single 2D gait template, avoiding matching features in temporal sequences. 
\\
Given a pre-processed binary gait silhouette sequence $I_t(x,y)$ at time $t$, the GEI is computed as:
\begin{equation}
    G(x,y) = \frac{1}{N} \sum_{t=0}^{N-1} I_t(p,q)
\end{equation}
GEI is comparatively robust to noise, but loses dynamic variations between successive frames. In order to retain dynamic changes in gait information, Zhang, Zhao and Xiong \cite{14zhang2010} proposed an active energy image (AEI) method for gait recognition, which focuses more on dynamic regions than GEI, and alleviates the effect caused by low quality silhouettes. \par Current research on gait representation include Gait Entropy Image (GEnI) \cite{ahmedgait}, frequency-domain gait entropy (EnDFT) \cite{ahmedgait}, gait energy volume (GEV) etc., which focus more on dynamic areas and reducing view-dependence of traditional appearance-based techniques.

\newpage
\section{Proposed Methodology}

\par In this project, we proposed a motion-based approach to person identification, using \emph{affine moment invariants} (AMIs) as feature descriptors. \emph{Active Energy Image} (AEI) are generated from the subject silhouette sequence, which retains the dynamic variations in the gait cycle.

\par The approach can be summarized in the following steps:
\begin{enumerate}
    \item Extract AEI from the gait sequence and divide the image into multiple segments.
    \item Extract AMI from each segment to use as gait features. The database consists of affine moment invariants of multiple people who wear standard clothing with no accessories.
    \item AEI of test subject is also segmented like that of the dataset and AMIs are computed.
    \item Matching weights are estimated at each area based on similarity between features of the subject segments and the database.
    \item The subject is predicted  by the weighted integration of the similarities of all segments. k-Nearest Neighbor classifier is used to classify the test subject. k=1 for our experimental analysis.
\end{enumerate}

\subsection{Active Energy Image}
\par In gait recognition, two types of information - static and dynamic -  are extracted from the gait sequence. According to \cite{13han2006}, GEI is efficient in extracting static and dynamic information, both implicitly and explicitly. However, since GEI represents gait as a single image, there is a loss of dynamic information such as the fore-and-aft frame relations. Active Energy Image (AEI) feature representation can be used to solve the above problems.

\begin{figure}[!htb]
    \centering
    \includegraphics[width = 1.05\textwidth, height = 15mm]{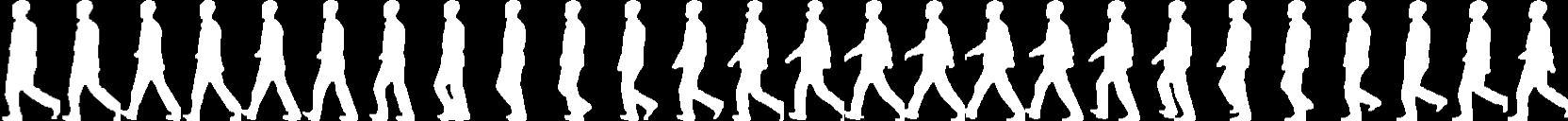}
    \caption{Gait silhouette sequence of individual}
    \label{fig:Gait Silhouette Sequence}
\end{figure}

\par
\begin{figure}[!htb]
    \centering
    \includegraphics[width = 1.05\textwidth, height = 15mm]{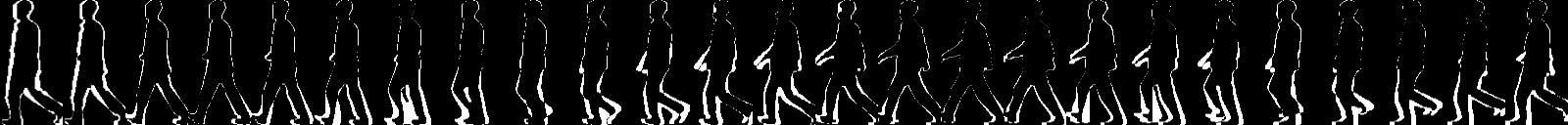}
    \caption{Difference Images between consecutive significant frames of gait sequence}
    \label{fig:Difference Images}
\end{figure}

 Given a pre-processed binary gait silhouette\cite{14zhang2010} $i = {i_0,i_1,....,i_N-1}$, where $f_j$ represents the j\textsuperscript{th} silhouette (Figure \ref{fig:Gait Silhouette Sequence}), N is the total number of frames in the sequence, the difference image between frames is calculated as follows (Figure \ref{fig:Difference Images}):

\begin{equation}
    I_j = \begin{cases} i_j(p,q) , \hfill j=0 \\ ||i_j(p,q) - i_{j-1}(p,q)||, \hfill j>0  \end{cases}
\end{equation}

\begin{figure}[!htb]
    \centering
    \includegraphics[height = 50mm]{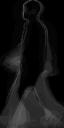}
    \caption{Active Energy Image}
    \label{fig:AEI}
\end{figure}
The AEI is defined as (Figure \ref{fig:AEI}):
\begin{equation}
    A(p,q) = \frac{1}{N} \sum_{j=0}^{N-1} I_j(p,q)
\end{equation}
where $N$ is the total number of difference images used to compute the AEI.

\par The AEI representation is divided into 'K' segments as shown in Figure {\ref{fig:segments}}:
\begin{figure}[H]
    \centering
    \includegraphics[scale=1.5]{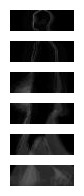}
    \caption{Segmented AEI (K = 6)}
    \label{fig:segments}
\end{figure}

\subsection{Affine Moment Invariants}
\textbf{Image moments and moment invariants}
\par Image moments are weighted averages of the image pixels' intensities or a function of such moments, usually chosen due to some attractive property or interpretation. Some properties of images derived through image moments include area, centroid, and information about the image orientation.
\par Image moments are used to derive \textbf{moment invariants}, which are invariant to transformations such as translation and scaling. The well-known \textbf{\emph{Hu moment invariants}} were shown to be invariant to translation, scale and rotation. However, Flusser\cite{29Flusser2000OnTI} showed that the traditional set of Hu moment invariants is neither independent nor complete.

\par \emph{Affine moment invariants} (\emph{AMIs}), proposed by Flusser and Suk \cite{17flusser1993} are moment-based descriptors, which are invariant under a general affine transformation. The invariants are generally derived by means of the classical theory of algebraic invariants\cite{18hilbert1993}, graph theory, or the method of normalization. The most common method is the use of graph theory.

\par The moments describe shape properties of an object as it appears. For an image, the centralized moment of order \emph{(a+b)} of an object \emph{O} is given by
\begin{equation}
    \mu_{ab} = \sum\sum_{(x,y)\in O} (x-x_{cg})^{a} (y-y_{cg})^{b} A(x,y)
\end{equation}
Here, $x_{cg}$ and $y_{cg}$ define the center of the object, calculated from the geometric moments $m_{ab}$, given by $x_{cg} =\frac{m_{10}}{m_{00}}$ and $y_{cg} =\frac{m_{01}}{m_{00}}$.

The affine transformation is expressed as
\begin{equation}
    u = a_0 + a_{1}x+a_{2}y
\end{equation}
\begin{equation}
    v = b_0 + b_{1}x+b_{2}y
\end{equation}

\cite{20suk2005} We have used 10 AMIs ($\textbf{A} = (A_1,A_2,....,A_{10})^T$), 5 of which are shown below:
\begin{figure}[H]
    \centering
    \includegraphics[width = 0.75\textwidth]{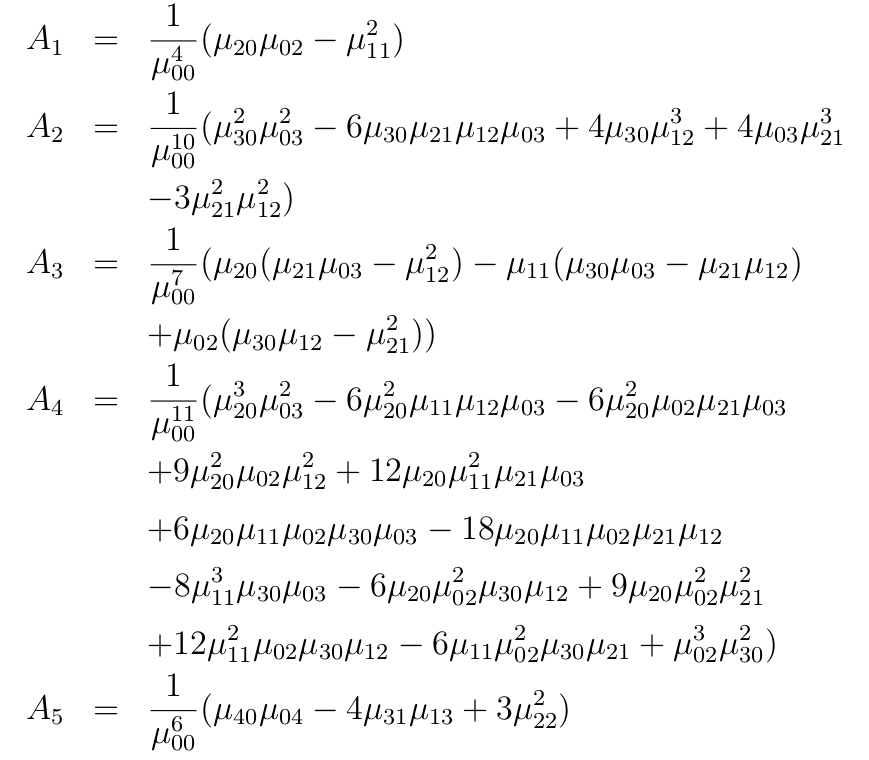}
    \label{fig:ami}
\end{figure}

\subsection{Estimation of matching weights}
\par \cite{05iwashita2013} Matching weights are calculated by first whitening the AMI in the database and the subject for each segment. $d_{n,s}^{k}$, which is the L2 norm between the unknown subject features and those of all known persons in the database, is computed as follows:
\begin{equation}
    d_{n,s}^{k} = || ^{w}A_{SUB}^k - ^{w}A_{DB_{n,s}}^k ||
\end{equation}
where $^{w}A_{SUB}^k$ and $^{w}A_{DB_{n,s}}^k$ are the whitened AMI of the test subject and of a known person  in the database respectively. \\ (\emph{n,s and k}) are $1 \leq n \leq N$ (N is number of persons in database), $1 \leq s \leq S$ (S is number of sequences of each person), and $1 \leq k \leq K$ (K is number of divided areas). $d_{n,s}^k$ is calculated in the Euclidean norm.
\\
\par The AMIs are whitened by:
\begin{enumerate}
    \item Applying \emph{principal component analysis (PCA)} on the AMIs and projecting them to a new feature space.
    \item Normalizing the projected features based on the corresponding eigenvalues.
\end{enumerate}

\par To estimate matching weights, we use the similarity between subject features and database features; high matching weights are set to areas with less appearance changes, and low matching weights set to those with more appearance changes.
\newline
\par Steps to estimate matching weights:
\begin{enumerate}
    \item At each segment k, select sequences where $d_{n,s}^k < \Bar{d}_{min}$. These selected sequences are considered to have high similarity with the subject. $\Bar{d}_{min}$ is defined as:
    \begin{equation}
        \Bar{d}_{min} = min_n \Bar{d}_n^k 
    \end{equation}
    \begin{equation}
        \Bar{d}_n^k = \dfrac{1}{S}\sum_{s=1}^{S} d_{n,s}^k
    \end{equation}
    where $S$ is the total number of sequences of the person being used. We consider that in each segment, if a minimum of one sequence of a person in the database is selected, then the matching scores of all sequences of that person are also high.
    \item The non-selected sequences are all considered to be low similarities, so the distances of these sequences are redefined as \emph{$d_{max}$ ($d_{max} = max_{n,s,k} d_{n,s}^k$)}
    \item These steps are applied for all segments and the total distance for all segments is calculated by $D_{n,s} = \sum_{k=1}^{K} d_{n,s}^k$. Subject is identified by nearest-neighbor method.
\end{enumerate}

\subsection{Principal Component Analysis}
\par The curse of dimensionality shows that as the number of features increase, it gets harder to visualize the training set and then work on it. Sometimes, many of these features are correlated, and hence redundant. This leads to a need for reducing the complexity of a model to avoid overfitting.\par \emph{Principal Component Analysis (PCA)} \cite{Jolliffe2002Principal} is a popular algorithm used for dimension reduction. Proposed by Karl Pearson, PCA is an unsupervised linear transformation technique used in identifying patterns in data based on the correlation in features. It projects the data onto a new subspace along the direction of increasing variance with equal or lesser dimensions.
\par The PCA Algorithm consists of the following steps:
\begin{enumerate}
    \item First the mean vector is computed.
    \item Assemble all the data samples in a mean adjusted matrix, followed by creation of the covariance matrix.
    \item Compute the Eigenvectors and Eigenvalues following the basis vectors.
    \item Each sample is represented as a linear combination of basis vectors.
\end{enumerate}

\par
\begin{figure}
    \centering
    \includegraphics[width=\textwidth]{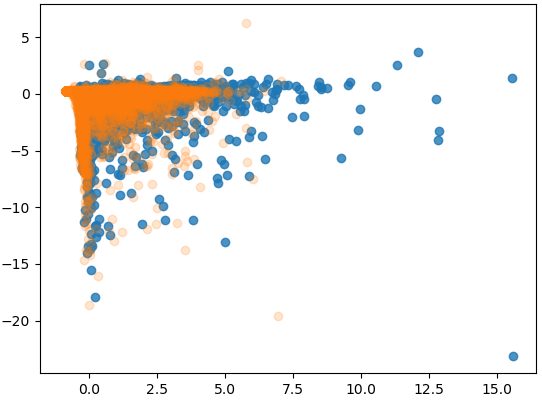}
    \caption{PCA on gait data extracted from CASIA-B Dataset. Blue datapoints represent original image data and yellow datapoints represent the modified data after dimensionality reduction.}
    \label{fig:my_label}
\end{figure}

 Figure \ref{fig:my_label} compares the distribution of image data before and after PCA dimensionality reduction.

\newpage
\subsection{k-Nearest Neighbor Classifier}
The k-Nearest Neighbor (kNN) algorithm is one of the simplest Machine Learning algorithms based on Supervised Learning technique. It calculates the similarity between the new case/data and available cases and classifies the new case into the category closest to the existing classes. 
\\ The kNN algorithm stores the dataset during the training phase; during the testing phase, when it gets new data, it classifies that data into a category that is much similar to the new data. The key steps are as follows:
\begin{enumerate}
    \item The number of neighbors (k) are selected.
    \item The Euclidean distance of k number of neighbors is calculated.
    \item The k nearest neighbors as per the calculated Euclidean distance are selected.
    \item Among these k neighbors, the number of the data points in each category are counted.
    \item The new data points are assigned to that category for which the number of neighbors is maximum.
\end{enumerate}

\section{Experimental Results \& Observations}

\par Our proposed method was applied to the CASIA-B Gait Dataset. A large multi-view gait database created in January 2005, it contains 124 subjects captured from 11 views. Three variations, namely view angle, clothing and carrying condition changes, are separately considered. 

\par In the experiment, we used the lateral view (90 degree) standard walking sequences to analyze the prediction rate of the algorithm. The CASIA-B Dataset consists of six standard walking sequences for each person. CCRs are calculated by dividing the six sequences of each subject into two sets; 124 x \emph{S} sequences were used for training (S = 3,4,5) and the rest were used for testing. All input images from the sequences are resized to 128 x 64 images for uniformity.

\par The total number of AMIs M were varied between 1 to 10 using PCA dimensionality reduction and the parameter K (number of divided areas) from 5 to 30. In case of $K = 23$ and $M = 5$, the proposed method shows the highest accuracy of $\mathbf{91.13}\%$ (Table \ref{table:results}). It is observed that increasing the number of image segments over a certain limit reduces the accuracy of the classification algorithm. This is likely due to the loss of features as the number of segments increases.

\begin{table}[h!]
    \centering
    \caption{Experimental results of our algorithm on the CASIA-B Gait Dataset}
    \begin{tabular}{| c | c | c |} 
 \hline
 \textbf{Train/Test Split} & \textbf{No. of Segments} & \textbf{Classification Accuracy} \\ [0.5ex] 
 \hline
 \multirow{4}{3em}{0.5} & 10 & 71\% \\
 & 20 & 78.76\% \\
 & 23 & \textbf{82.00\%} \\
 & 30 & 79.57\% \\
 \hline
 \multirow{4}{3em}{0.66} & 10 & 76.61\% \\
 & 20 & 85.08\% \\
 & 23 & \textbf{85.89\%} \\
 & 30 & 74.68\% \\
 \hline
 \multirow{4}{3em}{0.83} & 10 & 82.26\% \\
 & 20 & 90.32\% \\
 & 23 & \textbf{91.13\%} \\
 & 30 & 87.90\% \\
 \hline
\end{tabular}
    \label{table:results}
\end{table}

\par We also compared our proposed algorithm with existing methodologies that have been tested on the CASIA-B Gait Dataset, shown in Table \ref{tab:comparison}. Our approach is competitive, which shows the effectiveness of evaluating dynamic variations for recognition.
\begin{table}[h!]
    \centering
    \caption{Comparision of Classification Accuracy on CASIA-B Gait Dataset}
    \begin{tabular}{ | c | c | } 
    \hline
    \textbf{Method} & \textbf{Rank - 1 Classification} \\
    \hline
    Method 1 \cite{05iwashita2013} & 97.70\% \\ \hline
    Method 2 \cite{2016zhiyongAEI} & 95.52\% \\ \hline
    AEI+2DLPP \cite{14zhang2010} & 98.39\% \\ \hline
    STIPs+BoW \cite{2014WorapanIVC} & 94.5\% \\\hline
    Proposed Method & 91.13\%\\
    \hline
    \end{tabular}
    \label{tab:comparison}
\end{table}
\newpage
\section{Conclusion}
This paper describes a modified approach to gait-based person re-identification using AEIs. AEIs were generated to locate and identify the dynamic motion regions in a walking sequence, which were then resized and segmented laterally. Then we extracted AMIs as descriptors from each segment, applying PCA analysis to reduce the dimensionality and maximize variance in the feature space. The reduced features were assigned matching weights based on the similarity between the test subjects and the database, which were then classified using the combination of similarities in the segments. The experimental results demonstrate that our proposed method provides results which rival that of other gait recognition methods. Our future work will focus on improving the performance of our algorithm and evaluating its performance on walking sequences from arbitrary viewpoints.
%
%
%
%
%
%
\label{Bibliography}
\bibliographystyle{splncs}
\nocite{*}
\bibliography{bibliography.bib}

\begin{thebibliography}{10}

\bibitem{11sarkar2005}
Sarkar, S., Phillips, P.J., Liu, Z., Robledo~Vega, I., Grother, P., Bowyer, K.:
\newblock The humanid gait challenge problem: Data sets, performance, and
  analysis.
\newblock IEEE transactions on pattern analysis and machine intelligence
  \textbf{27} (03 2005)  162--77

\bibitem{Carley2019PersonRF}
Carley, C., Ristani, E., Tomasi, C.:
\newblock Person re-identification from gait using an autocorrelation network.
\newblock In: CVPR Workshops. (2019)

\bibitem{12bobick2001}
Bobick, A., Johnson, A.:
\newblock Gait recognition using static, activity-specific parameters.
\newblock Volume~1. (02 2001)  I--423

\bibitem{kolekar2018intelligent}
Kolekar, M., Francis, T..:
\newblock Intelligent Video Surveillance Systems: An Algorithmic Approach.
\newblock Chapman and Hall/CRC (2018)

\bibitem{13han2006}
Han, J., Bhanu, B.:
\newblock "individual recognition using gait energy image".
\newblock IEEE transactions on pattern analysis and machine intelligence
  \textbf{28} (03 2006)  316--22

\bibitem{14zhang2010}
Zhang, E., Zhao, Y., Xiong, W.:
\newblock Active energy image plus 2dlpp for gait recognition.
\newblock Signal Processing \textbf{90} (07 2010)  2295--2302

\bibitem{ahmedgait}
Ahmed, I., Rokanujjaman, M.:
\newblock Gait-based person identification considering clothing variation.
\newblock \textbf{5} (06 2016)  28--42

\bibitem{29Flusser2000OnTI}
Flusser, J.:
\newblock On the independence of rotation moment invariants.
\newblock Pattern Recognition \textbf{33} (2000)  1405--1410

\bibitem{17flusser1993}
Flusser, J., Suk, T.:
\newblock Pattern recognition by affine moment invariants.
\newblock Pattern Recognition \textbf{26} (1993)  167--174

\bibitem{18hilbert1993}
Marxsen, S., Hilbert, D., Laubenbacher, R., Sturmfels, B., David, H.,
  Laubenbacher, R.:
\newblock Theory of Algebraic Invariants.
\newblock Cambridge Mathematical Library. Cambridge University Press (1993)

\bibitem{20suk2005}
Suk, T.:
\newblock Tables of affine moment invariants generated by the graph method.
\newblock (01 2005)

\bibitem{05iwashita2013}
Iwashita, Y., Uchino, K., Kurazume, R.:
\newblock Gait-based person identification robust to changes in appearance.
\newblock Sensors (Basel, Switzerland) \textbf{13} (06 2013)  7884--901

\bibitem{Jolliffe2002Principal}
Jolliffe, I.T.:
\newblock Principal Component Analysis.
\newblock Springer Series in Statistics. Springer-Verlag, New York (2002)

\bibitem{2016zhiyongAEI}
Liu, Z.:
\newblock Gait recognition using active energy image and gabor wavelet.
\newblock (10 2016)  1354--1358

\bibitem{2014WorapanIVC}
Kusakunniran, W.:
\newblock Attribute-based learning for gait recognition using spatio-temporal
  interest points.
\newblock Image and Vision Computing \textbf{32} (11 2014)

\bibitem{01gala2014}
Gala, A., Shah, S.:
\newblock Gait-assisted person re-identification in wide area surveillance.
\newblock (04 2015)  633--649

\bibitem{19flusser2004}
Suk, T., Flusser, J.:
\newblock Graph method for generating affine moment invariants.
\newblock Volume~2. (09 2004)  192 -- 195 Vol.2

\end{thebibliography}





\end{document}